\documentclass[letterpaper, 10 pt,  journal, twoside]{ieeetran}  

\renewcommand{\thefootnote}{\fnsymbol{footnote}}

\DeclareRobustCommand{\commonfignote}{%
  \texorpdfstring{\hyperlink{fn:commonfig}{\textsuperscript{$\S$}}}{}%
}

\usepackage{cite}
\usepackage{amsmath,amssymb,amsfonts}
\usepackage{algorithmic}
\usepackage{graphicx}
\usepackage{textcomp}
\usepackage{xcolor}
\usepackage{float}
\usepackage{caption}
\usepackage{scalerel}
\usepackage{bm}
\usepackage{tikz}
\usepackage{hyperref}
\usepackage{booktabs}
\usepackage{enumitem}

\usepackage[dvipsnames]{xcolor}

\hypersetup{hidelinks}

\begin{document}

\markboth{IEEE Robotics and Automation Letters. Preprint Version. Accepted September, 2026}
{Delbene \MakeLowercase{\textit{et al.}}: Compact Force Sensor for Dual-UAV Cable-Suspended Payload Transport}

\author{Andrea Delbene$^{1}$, Giorgio Cannata$^{1}$, Giorgio Carlini$^{1}$, Filippo Sante$^{1}$, and Marco Baglietto$^{1}$
\thanks{Manuscript received: July, 16, 2026; Accepted September, 6, 2026.}
\thanks{This paper was recommended for publication by Editor G. Loianno upon evaluation of the Associate Editor and Reviewers’ comments.}
\thanks{$^{1}$ The authors are with the Department of Informatics, Bioengineering, Robotics and Systems Engineering, Università degli Studi di Genova, Via all'Opera Pia 13, 16145, Genoa, Italy {\tt\footnotesize andrea.delbene@edu.unige.it; giorgio.cannata@unige.it; giorgio.carlini@unige.it; filippo.sante@unige.it; marco.baglietto@unige.it}}
\thanks{Digital Object Identifier (DOI): see top of this page.}
}

\title{Compact Force Sensor for Dual-UAV Cable-Suspended Payload Transport with Tension-Aware Outer-Loop Control}

\maketitle

\begin{abstract}
Cooperative payload transportation using multiple \textit{Unmanned Aerial Vehicles} (UAVs) poses challenges in stability, coordination, and robustness, especially under external disturbances and unmodeled dynamics. This work proposes a dual-UAV payload transportation framework supported by a compact, custom-designed force sensor measuring the interaction force at the UAV cable anchor point. The sensor design and mathematical model are presented, and its performance is characterized through static and dynamic tests evaluating linearity, hysteresis, repeatability, and crossload. The control architecture follows a cascade structure: fast inner loops handle vehicle stabilization, while outer loops are designed to compensate for the measured forces. The approach is validated through simulations and indoor experiments under position uncertainty. Payload-drop and constrained-space tests assess the proposed sensing and control architecture against literature-based distributed references, showing improved stabilization, coordination, and disturbance rejection. A video of the experiments is available at: \url{https://youtu.be/rIw9-fvV8Qw}.
\end{abstract}
\vspace{0.2cm}
\begin{IEEEkeywords}
Aerial Systems: Mechanics and Control; Cooperating Robots; Distributed Robot Systems.
\end{IEEEkeywords}
\vspace{-0.3cm}
\section{Introduction} \label{Introduction}
\IEEEPARstart{I}{n} recent years, UAVs have become increasingly popular for cooperative tasks such as payload transportation, which introduces challenges including the design of the UAV-payload connection. Solutions include robotic arms \cite{CaccavaleRobArms}, rigid connections with grippers or magnets \cite{Tagliabue, Loianno}, and rigid bars \cite{Bosio}, each prioritizing structural rigidity over formation modularity. Flexible cables remain the most practical choice, enabling full payload pose control \cite{DarioFirstPP, PickPlaceDario} and formation changes for collision avoidance \cite{Jackson}.

Other challenges include the design of stable and robust decentralized control under limited or absent inter-UAV communication, and modeling, estimating, or measuring interaction forces. Approaches assuming full or partial state knowledge use \textit{Optimal} or \textit{Model Predictive Control} \cite{Jackson, MPCSundin, Buzzurro}, while leader-follower strategies are common with limited communication: in \cite{Scaramuzza}, authors use a \textit{Linear Quadratic Regulator} on both UAVs with only \textit{Inertial Measurement Unit} (IMU) and \textit{Visual Odometry}, whereas authors in \cite{XieLasVegas} neutralize leader accelerations from IMU data. Leader-follower admittance controllers have also been proposed for force compensation \cite{Tagliabue, GabellieriForceCtrl, TognonIF, TagliabueOptSens}.

\begin{figure}[t]
\centerline{\includegraphics[scale=0.11]{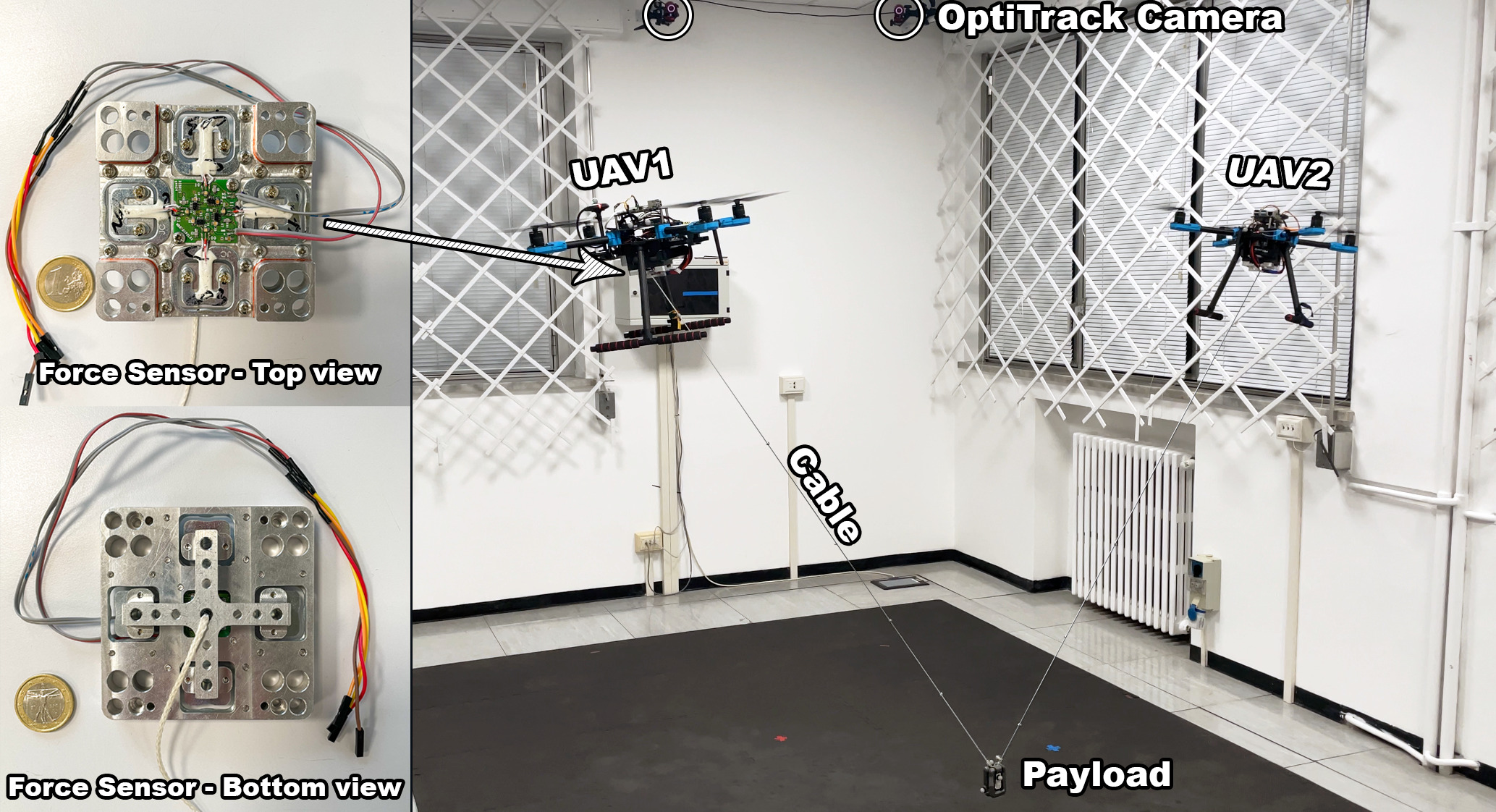}}
\caption{Experimental setup: two PX4/ROS2-powered UAVs connected to a 1 kg payload by nylon cables. Left column: top and bottom views of the custom force sensor.}
\vspace{-0.5cm}
\label{system_force_sensor}
\end{figure}

Accurate estimation or direct measurement of UAV–payload interaction forces is essential for robust cooperative transport. Some works treat tensions as external disturbances \cite{ZhangRAL}, while others use state-based models: when full state information is available, such as in simulations or with \textit{Motion Capture} (MoCap), tensions are modeled from the system state \cite{Jackson}, \cite{MPCSundin}, \cite{PereiraTensionState, HRCollabTransp, SunDario, DelbeneTension}, or computed from load statics and nullspace force allocation \cite{GabellieriCyclicNullspace}. In \cite{SunDario, DelbeneTension}, tensions are modeled through a \textit{mass-spring-damper} system, with \cite{DelbeneTension} identifying its parameters via \textit{Nonlinear Least Squares}; in \cite{SunDario}, this model supports the EKF-based planner, while onboard external forces are estimated from measured acceleration and thrust. A more detailed cable model uses successions of point-like mass-spring-damper elements \cite{GimenezUltra}. In \cite{WahbaIROS24}, reference tensions are computed from planned acceleration, orientation, and thrust.

When states are uncertain and must be estimated, cable tensions can be inferred using various methods: reversing torque dynamics \cite{XieLasVegas}, combining \textit{Recursive Least-Squares Approximation} with an \textit{Acceleration-Based Force Observer} \cite{BendingPayload}, stochastic filtering using \textit{Extended} or \textit{Unscented Kalman Filters} \cite{Tagliabue, TagliabueOptSens, HRCollabTransp, SunDario}, or \textit{Momentum-Based Observers} \cite{GabellieriForceCtrl, TognonIF, FlyCrane}. These approaches provide model-based, lightweight force predictions without additional hardware, but their accuracy depends on state estimation and model fidelity; moreover, these schemes typically require sufficiently informative measurements and communication to reconstruct the full system state, and they inherently introduce delays in response to abrupt force variations.

Such limitations are critical in applications requiring fast responses, such as \cite{DelbeneRecovery}, where the authors study the failure of a single UAV during transportation and propose rapid recovery strategies to avoid compromising the mission. Moreover, they are relevant in scenarios where full state information—especially the payload state—is not available, such as outdoor deployments without MoCap, and with potentially limited inter-UAV communication.

A practical alternative is then to employ commercial force sensors, as in \cite{EnergyDistr, BernardForceSens1, BernardForceSens2, SRIM37} (the latter considers a single-UAV setup). Unlike estimators, force sensors are model-independent, provide direct high-rate measurements, and can serve as ground-truth instruments in research. Drawbacks include added weight, integration complexity, and sensitivity to electronic noise and disturbances. In \cite{EnergyDistr, BernardForceSens1, BernardForceSens2}, a uniaxial force sensor is used, which is more affordable than the six-axis force/torque sensor adopted in \cite{SRIM37}; however, reconstructing three-axis forces requires extra hardware (e.g., cardan joints and magnetic encoders), making the system bulky and heavy for aerial platforms. For validation, \cite{HRCollabTransp} relies on a uniaxial load cell combined with MoCap, whereas \cite{TagliabueOptSens} employs a three-axis optical force sensor that is smaller and lighter but more delicate and typically less accurate than \textit{strain-gauge} or capacitive sensors.

The main contribution of this letter is the development of a custom, compact three-axis force sensor that provides a low-cost alternative to those commonly adopted in related work, while maintaining comparable performance, weight, and size. To validate the sensor, we consider a PX4/ROS2-based framework for cooperative payload transportation with two UAVs connected by flexible cables, in which the sensor integrates straightforwardly, and we introduce a decentralized cascade-control architecture explicitly accounting for cable tension. The framework requires limited information sharing: each UAV uses only its own state and exchanges minimal data for state transitions.

The work is organized as follows: Section~\ref{system_modeling} presents the mathematical model of the system, while Section~\ref{force_sensor_design} details the design and characterization of the custom force sensor, comparing it with sensors reported in the literature. Section~\ref{control_architecture} introduces the decentralized cascade-control strategy, and Section~\ref{experimental_setup} describes the software and hardware architecture and the evaluation method. Section~\ref{results} reports the simulation and indoor experimental results, including payload-drop tests and maneuvers in a constrained space. Finally, Section~\ref{Conclusions} discusses conclusions and future work.

\vspace{-0.1cm}
\section{System Modeling}\label{system_modeling}
Two quadrotors transporting a payload through flexible cables with equal lengths are considered, as shown in Fig.~\ref{system_force_sensor}. The attachment points for both quadrotors and the payload are placed with an offset from their \textit{center of mass} (COM).
This section briefly introduces the system dynamics. In the following, explicit time dependencies are omitted for conciseness, vectors are denoted in boldface to differentiate them from scalars and matrices, $i = 1,2$ identifies the $i$-th quadrotor, $p$ refers to the payload, and a superscript dot indicates time differentiation.

A reference inertial \textit{North-East-Down} (NED) frame is defined as $\mathcal{F}_I(\bm{O}_I, \bm{x}_I, \bm{y}_I, \bm{z}_I)$. Additionally, three local body-fixed frames are introduced: one for each quadrotor, $\mathcal{F}_i(\bm{O}_i, \bm{x}_i, \bm{y}_i, \bm{z}_i)$ and one for the payload $\mathcal{F}_p(\bm{O}_p, \bm{x}_p, \bm{y}_p, \bm{z}_p)$, with the $z$-axis pointing downwards. For the $i$-th UAV, the state vector is defined as $\bm{\nu}_i = [\bm{\xi}_i, \dot{\bm{\xi}}_i, \bm{\eta}_i, \dot{\bm{\eta}}_i]^\top = [x_i \, y_i \, z_i \, \dot{x}_i \, \dot{y}_i \, \dot{z}_i \, \phi_i \, \theta_i \, \psi_i \, \dot{\phi}_i \, \dot{\theta}_i \, \dot{\psi}_i]^\top \in \mathbb{R}^{12}$ with an analogous definition $\bm{\nu}_p$ for the payload. There, $\bm{\eta}_i = [\phi_i \, \theta_i \, \psi_i]^\top$ indicates the $i$-th UAV orientation expressed with \textit{Tait-Bryan} angles (roll, pitch, yaw), while $\dot{\bm{\eta}}_i$ the angular velocities; same is valid for the payload. The translational dynamics of the UAVs and payload, derived from the classical Newton-Euler formulation, are expressed as:
\begin{equation}
    \begin{cases}
    m_i \ddot{\bm{\xi}}_i = -u_i R_i \bm{e}_3 + \bm{P}_i + \bm{T}_i\\
    m_p \ddot{\bm{\xi}}_p = \sum_{i=1}^{2}\bm{T}_{ip} + \bm{P}_p 
    \end{cases}\quad , \label{trans_eqs}
\end{equation}
while the rotational dynamics are given by:
\begin{equation}
    \begin{cases}
            I_i \dot{\bm{\omega}}_i = \bm{\tau}_i + \bm{r}_i \times (R_i^\top \bm{T}_i) - \bm{\omega}_i \times (I_i \bm{\omega}_i) \\
            I_p \dot{\bm{\omega}}_p = \sum_{i=1}^{2}\bm{r}_{ip} \times (R_p^\top \bm{T}_{ip}) - \bm{\omega}_p \times (I_p \bm{\omega}_p) 
    \end{cases}\quad . \label{rot_eqs}
\end{equation}
Here, $R_i, R_p \in \mathbb{R}^{3 \times 3}$ are rotation matrices mapping body-fixed frames to the inertial frame $\mathcal{F}_I$. For the $i$-th UAV, $u_i$ represents the thrust generated, with $R_i \bm{e}_3$ $(\bm{e}_3 = [0 \, 0 \, 1]^\top$) expressing it in $\mathcal{F}_I$; $\bm{\tau}_i$ is instead the control torque; $\bm{P}_i, \bm{P}_p$ are gravitational forces acting on the UAVs and payload, with masses $m_i$ and $m_p$. Cartesian positions are denoted as $\bm{\xi}_i, \bm{\xi}_p \in \mathbb{R}^3$, while $\dot{\bm{\xi}}_i, \dot{\bm{\xi}}_p$ and $\ddot{\bm{\xi}}_i, \ddot{\bm{\xi}}_p$ correspond to linear velocities and accelerations, respectively; body angular velocities and accelerations are denoted as $\bm{\omega}_i, \bm{\omega}_p$ and $\dot{\bm{\omega}}_i, \dot{\bm{\omega}}_p$. The inertia matrices are diagonal and are indicated as $I_i, I_p \in \mathbb{R}^{3 \times 3}$. The vectors $\bm{r}_i, \bm{r}_{i,p}$ define the cable anchor-point locations wrt the $i$-th UAV and payload COM, respectively. Finally, the cable tension acting on the $i$-th UAV and the payload are $\bm{T}_i = [T_{i,x} \; T_{i,y} \; T_{i,z}]^\top$ and $\bm{T}_{ip} = [T_{ip,x} \; T_{ip,y} \; T_{ip,z}]^\top$, respectively, expressed in $\mathcal{F}_I$.

\section{Force Sensor Design}\label{force_sensor_design}
\subsection{Sensor Description}
\begin{figure}[t!]
\vspace{0.1cm}
\centerline{\includegraphics[scale=0.076]{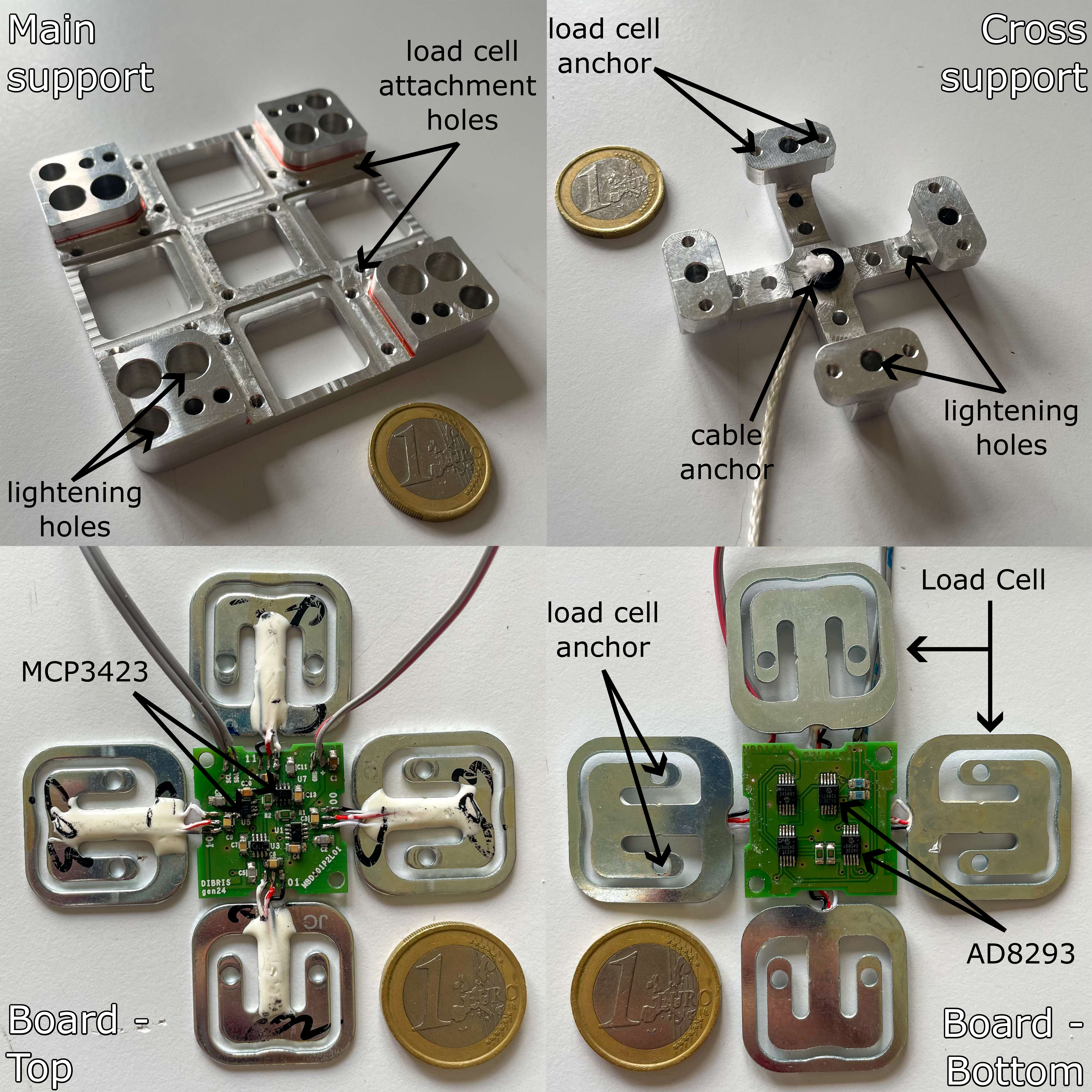}}
\caption{Force sensor components: top-left, base supporting the cells and board; top-right, cross structure linking the cells with the anchor point; bottom-left and bottom-right, load cells and electronic board (top and bottom views).}
\vspace{-0.4cm}
\label{force_sensor_all}
\end{figure}

A custom strain-gauge-based force sensor was designed to measure cable tension. It consists of four small load cells arranged in a Wheatstone bridge configuration and integrated into a compact mechanical structure. Each cell supports up to $5$~kg, includes two $1$~k$\Omega$ active gauges, has a precision of $0.05$\% Full Scale (FS), and an overload capacity of $120$\%~FS. As shown in Fig.~\ref{force_sensor_all}, the load cells are mounted on an aluminium base and connected to a dedicated \textit{Printed Circuit Board} (PCB), while a cross-shaped support links the UAV to the payload, with each arm rigidly attached to a load cell.

The compact PCB is centrally housed within the load-cell arrangement to minimize connection length and electromagnetic interference, while remaining securely attached to the mechanical support. It conditions and digitizes the load-cell signals through four parallel \textbf{MCP3423} sigma-delta \textit{analog-to-digital} converters, each operating at $240$~\textit{samples per second} with $12$-bit resolution. The converters share a single I$^2$C bus using unique chip addresses; this solution was preferred to a single multi-channel converter to maximize sampling frequency, despite increased power consumption and footprint.

As the MCP3423 internal amplifier gain is insufficient for the load cell signals, four \textbf{AD8293} differential amplifiers with a fixed gain of $160$ were added, acting also as anti-aliasing filters. A $0.5$~V reference, corresponding to the at-rest output of the load cells, is applied to the negative differential inputs and summed with the output, ensuring a positive signal compatible with a single power supply circuit.

\subsection{Measures to Force}
\begin{figure}[b]
\vspace{-0.3cm}
\centerline{\includegraphics[scale=0.305]{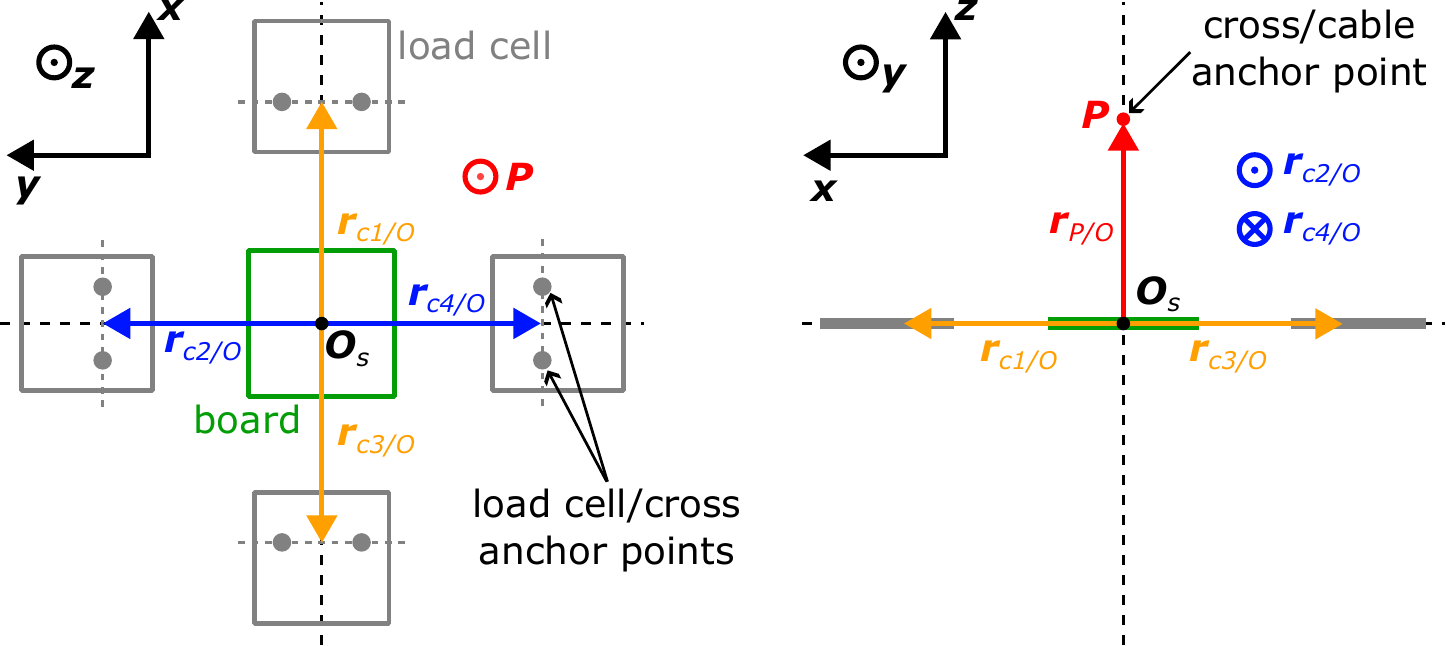}}
\caption{Geometric representation of load cells positions and cable anchor point relative to board center, in bottom (z-out) and lateral (y-out) views.}
\vspace{-0.3cm}
\label{load_cells_math_model}
\end{figure}

Measurements from individual load cells alone are not sufficient to determine the full three-dimensional force applied to the sensor. In this section, we explain how the forces measured by each cell are transformed into the force vector applied at the cable anchor point.
First, consider the sensor flipped, as shown in Fig.~\ref{load_cells_math_model}, and a local frame $\mathcal{F}_s$ centered at the origin $\bm{O}_s$ of the circuit board.
Each cell is located at position $\bm{c}_j, \;j = 1,2,3,4$, at a distance $d_0$ from $\bm{O}_s$, with vector position $\bm{r}_{c_j/O} = \bm{c}_j - \bm{O}_s$.
The anchor point $\bm{P}$ is at an altitude $h$ from $\bm{O}_s$, defined by $\bm{r}_{P/O} = \bm{P} - \bm{O}_s$.
Each cell measures, at position $\bm{c}_j$, a scalar force $f_{c_j} < 0$, indicating that the cell operates under tension.
The system is then supposed to be in static equilibrium\footnote{The displacement of the load cells is considered negligible.}, or:
\begin{equation}
    \begin{cases}
        {}^\mathcal{F}\bm{T} + \sum_{j=1}^{4} \bm{F}_{c_j} \; = \; \bm{0} \\
        \bm{r}_{P/O} \times {}^\mathcal{F}\bm{T} + \sum_{j=1}^{4}\bm{r}_{c_j/O} \times \bm{F}_{c_j} \; = \; \bm{0}
    \end{cases} \quad ,\label{static_equilibrium}
\end{equation}
where $\bm{F}_{c_j} = [0 \; 0 \;f_{c_j}]^\top$, is the vectorial definition of the force measured by cell $j$, and ${}^\mathcal{F}\bm{T} = [{}^\mathcal{F}T_x, {}^\mathcal{F}T_y, {}^\mathcal{F}T_z]^\top \in \mathbb{R}^{3}$ is the tension force in the sensor (or UAV) body frame. 
From the first equation, the vertical tension component follows:
\begin{equation}
    {}^\mathcal{F}T_z = -\sum_{j=1}^{4}f_{c_j} \quad .
\end{equation}\label{T_z}
Since the load cells measure only normal forces, the first relation in \eqref{static_equilibrium} is used only to recover ${}^\mathcal{F}T_z$; the lateral components ${}^\mathcal{F}T_x$ and ${}^\mathcal{F}T_y$ cannot be obtained from the direct force balance.
By then taking advantage of the symmetry $\bm{r}_{c_1/O} = - \bm{r}_{c_3/O}$ and $\bm{r}_{c_2/O} = - \bm{r}_{c_4/O}$, the moment equilibrium equation in \eqref{static_equilibrium} is developed: 
\begin{equation}
    \bm{r}_{P/O} \times {}^\mathcal{F}\bm{T} + \bm{r}_{c_1/O} \times (\bm{F}_{c_1} - \bm{F}_{c_3}) + \bm{r}_{c_2/O} \times (\bm{F}_{c_2} - \bm{F}_{c_4}) = \bm{0} \quad . \label{moments_static_developed}
\end{equation}
From equation \eqref{moments_static_developed}, it follows:
\begin{equation}
    {}^\mathcal{F}T_x = -\dfrac{d_0}{h}(f_{c_1} - f_{c_3}) \;,\; {}^\mathcal{F}T_y = -\dfrac{d_0}{h}(f_{c_2} - f_{c_4}) \; .
\end{equation}

The tension vector ${}^\mathcal{F}\bm{T}$ can then be reconstructed from the load cell measurements.
These computations are based on the assumption that the measurement retrieved from each load cell directly represents the true force applied to it.
In practice, the load cells must be calibrated through a series of static and dynamic tests. The next subsection shows the methods considered for this purpose and the respective results.
\vspace{-0.1cm}
\renewcommand{\arraystretch}{1.3} 
\begin{table*}[ht!] 
\vspace{0.21cm}
\centering
\begin{tabular}{|c|c|c|c|c|c|c|}
\hline
  & \textcolor{gray}{DDE-100N-002-000} & \textcolor{brown}{FRC4115-0} \cite{HRCollabTransp} & \textcolor{orange}{FSE1001} \cite{EnergyDistr} & \textcolor{blue}{OMD-20-FE-200N} \cite{TagliabueOptSens} & \textcolor{purple}{SRI-M3713A1} \cite{SRIM37} & \color{ForestGreen}{Custom Sensor} \\
\hline
Technology & Strain Gauge & Strain Gauge & Capacitive & Optical & Strain Gauge & Strain Gauge \\
\hline
Measured Axes & Uniaxial & Uniaxial & Uniaxial & 3-Axis Force & 6-Axis Force/Torque & 3-Axis Force \\
\hline
Dimension (mm) & Ø$32\times50$h  & $58$w$\times13$l$\times13$h & Ø$60\times16$h & $11$w$\times15$l$\times10$h & Ø$45\times19$h & $80$w$\times80$l$\times21$h \\
\hline
Mass (g)& $96$ & \textbf{N/A} ($\sim100$) & $80$ & $1.5$ & $106$ & $118$ \\
\hline
Excitation (V) & $2$-$15$ AC/DC & $5$ DC & $2.8$-$5.3$ DC & \textbf{N/A} ($0.24$W) & $5$ DC & $5$ DC \\
\hline
Data Rate (Hz) & $530$ & $50$ & $200$ & $1000$ & $2000$ & $240$ \\
\hline
OLC (\%FS) & $150$ & $120$ & $200$ & $200$ & $300$ & $450$ \\
\hline
Hyst. (\%FS) & $<0.3$ & $<0.3$ & \textbf{N/A} & $< 2$ & $< 0.5$ & $< 0.5$ \\
\hline
Non-Lin. (\%FS) & $<0.3$ & $<0.3$ & \textbf{N/A} & $< 2$ & $< 0.5$ & $< 0.5$ \\
\hline
Repeat. (\%FS) & $< 0.2$ & $<0.3$ & \textbf{N/A} & \textbf{N/A} & \textbf{N/A} & $< 0.2$ \\
\hline
XLoad (\%FS) & \textbf{N/A} & \textbf{N/A} & \textbf{N/A} & $< 5$ & $< 2$ & $< 0.6$ \\
\hline
Cost & Medium & Cheap & Medium & Medium-High & High & Cheap \\
\hline
Interface & Analog, RS-232 & USB & USB & USB, CAN, UART & Ethernet, RS-232, ... & I²C \\
\hline
\end{tabular}
\caption{Comparison with commercial sensors used in similar works in the literature. For sensors used in \cite{BernardForceSens1, BernardForceSens2}, exact models are unknown; data from a similar one (DDE-100N-002-000) are shown for reference. Our custom sensor has competitive accuracy and low cost; its size is slightly larger, and its acquisition rate is lower than most of the other devices.}\label{sensors_comparison}
\vspace{-0.4cm}
\end{table*}

\subsection{Calibration}
Calibration steps are performed in static and dynamic conditions. In static calibration, the sensor is fixed to a leveled structure and the mean of the first $3000$ raw force outputs is recorded for each cell $j$ ($f_{c_j,calib}$) under an applied force reference $f_{truth}$. Ideally, each of the four cells should measure $f_{truth}/4$, but structural imperfections cause slight deviations.
Raw force signals $f_{c_j,raw}$ are then converted as:
\begin{equation}
    {f}_{c_j} = \gamma_{c_j}\; f_{c_j,raw} \quad , \gamma_{c_j} = f_{c_j,calib}/f_{truth} \quad.
\end{equation}
To reduce high-frequency noise, a \textit{moving average filter} with a window of $4$ samples is applied, introducing a small time delay ($\sim6$~ms), acceptable for the $100$~Hz outer control loop.

\begin{figure}[b!]
\vspace{-0.2cm}
\centerline{\includegraphics[scale=0.5]{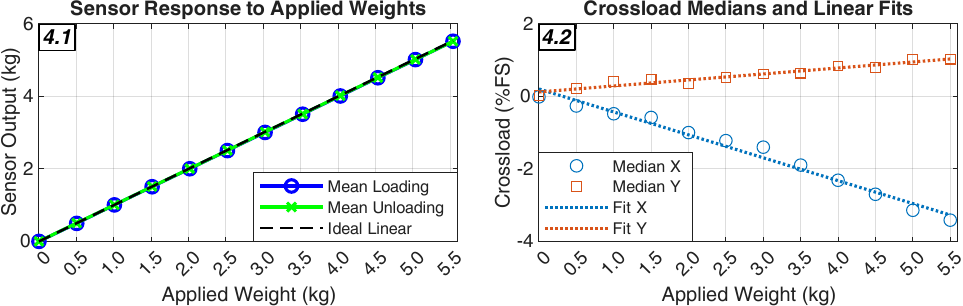}}
\caption{Left: Mean load/unload cycles per weight. Right: Median crossload per applied weight; dotted lines show linear fits illustrating growth with load. (tests on $z$-axis.)}
\vspace{-0.3cm}
\label{sensor_response_xload}
\end{figure}

Sensor performance is characterized using incremental weights up to $5.5$~kg applied on the $z$-axis, during $10$ loading/unloading cycles.
The following metrics are computed following practical calibration guidelines: \footnote{ISOBudgets,  sensor calibration resources, available at: \url{https://www.isobudgets.com/blog/}.} \textbf{hysteresis} (max difference between loading and unloading), \textbf{non-linearity} (max deviation from a linear fit), \textbf{repeatability} (consistency under repeated loads), and \textbf{crossload} (sensitivity on non-measurement axes to forces applied on another axis).
An initial analysis yielded a non-linearity of $1$\%~FS (Fig.~\ref{sensor_response_xload}.1); after applying a linear regression correction, this dropped to $0.5$\%~FS, while the others changed only slightly.

\begin{figure*}[t]
\vspace{0.2cm}
\centerline{\includegraphics[scale=0.46]{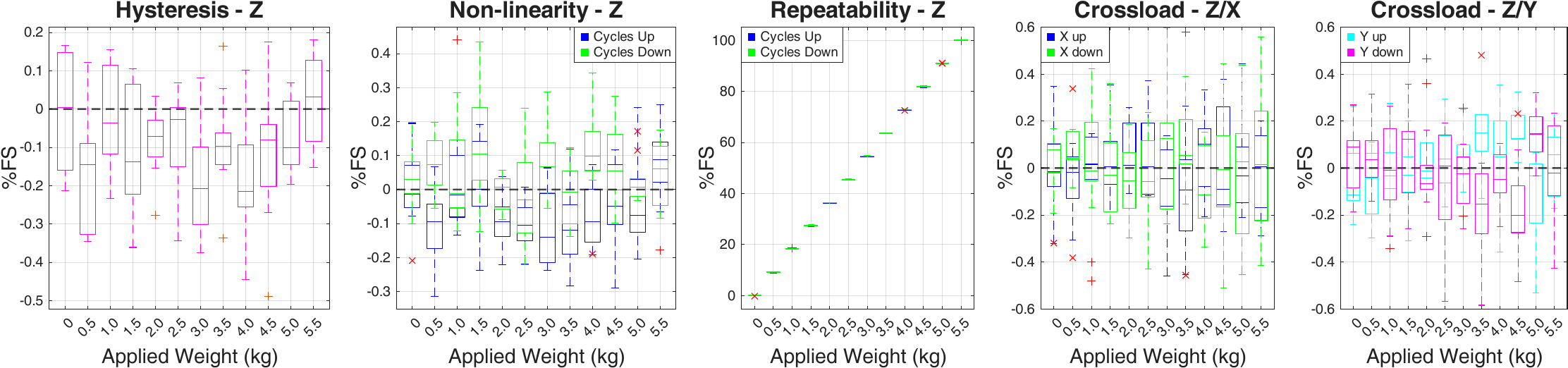}}
\caption{Box plots of Hysteresis, Non-Linearity, Repeatability and Crossload (\% FS) for each applied weight, relative to the measures taken on the $z$-axis test. Box body spans from first quartile (Q1 - lower side of the box) to third one (Q3 - upper side), horizontal line marks the median (Q2), and wicks show the observed minimum and maximum. Outliers are red $+$ for cycles up, $\times$ for down (indifferent in Hysteresis chart).}\label{hyst_nl_repeat_xload}
\vspace{-0.4cm}
\end{figure*}

Fig.~\ref{hyst_nl_repeat_xload} details the performance indexes. As shown in Table~\ref{sensors_comparison}, hysteresis and non-linearity are comparable with other sensors, while repeatability matches the DDE-100N-002-000. Crossload initially reached $3.8$\%~FS (Fig.~\ref{sensor_response_xload}.2); after applying a calibration-based matrix compensation on all axes, it drops to $0.6$\%~FS, excellent compared to alternatives. Improved mechanical alignment could achieve similar or better results.

The same testing procedure is repeated on the $x$ and $y$ axes to evaluate the sensor around its mechanical singularity, using the same metrics. These tests revealed a slight non-linear response that required higher-order regression, and their performance indexes were approximately twice those on the $z$-axis, confirming consistent performance even in less favorable orientations. However, its use in this work is far from these conditions; future work includes defining valid operating thresholds and/or adding a regression-switching rule when the measured forces get close to the singularities.

Dimensions and mass are higher, though the aluminum frame supporting the cells has a lot of room for optimization. The acquisition rate is lower than most of them, but it remains suitable for the target application. The overload capacity is the maximum load of the four cells, scaled to the considered FS. Unlike other sensors requiring USB, CAN, or Ethernet hardware/drivers, our sensor uses I²C and is read directly by a custom Python/ROS2 script. Considering both component and labor costs, the proposed sensor stands as a valid alternative to commercial solutions, enabling straightforward integration in UAV applications.

The dynamic behavior is tested using a practical alternative, as no setup was available to generate a three-axis step force. The sensor was mounted on the UAV inside a MoCap-equipped room, where fused MoCap-onboard IMU data provided accurate orientation. The UAV was manually tilted across roll and pitch angles up to $\sim\pm70$~°, while applying weights and recording three-axis tension and orientation. The expected force, computed from the applied weight and measured orientation, showed a linear relationship with the sensor output, aside from small scaling factors.
Although not a precise step-force test,  this method allows assessment of orientation-dependent behavior under realistic conditions. Instrumentation generating a controlled step force remains preferable for precise dynamic characterization.

\section{Control Architecture} \label{control_architecture}
\begin{figure}[!b]
\vspace{-0.2cm}
\centerline{\includegraphics[scale=0.47]{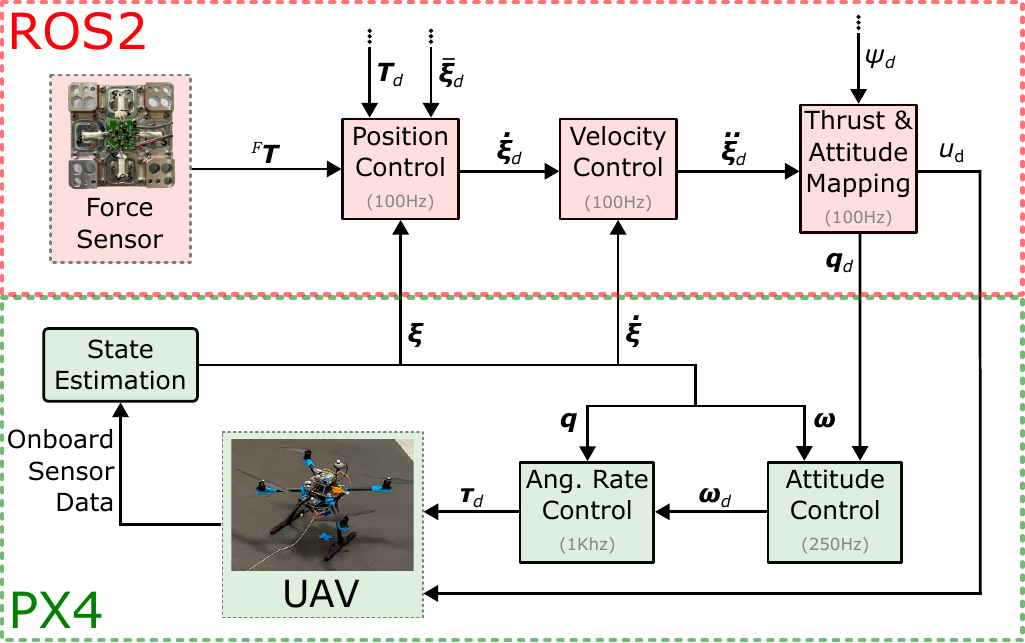}}
\caption{Cascaded distributed control scheme for each UAV. Red-dotted blocks run on ROS2, while green-dotted blocks are handled by PX4. Desired position, yaw, and cable tension generate quaternion and thrust references for the inner loops.}
\vspace{-0.3cm}
\label{cascaded_control}
\end{figure}

The control strategy is decentralized and organized as a cascade architecture inspired by \cite{PX4_ICRA}, where each high-level block generates references for the subsequent lower-level blocks (see Fig.~\ref{cascaded_control}). Each UAV computes its own references using only onboard estimates, without payload or other UAV data. The objective is to demonstrate how the custom force sensor improves in-flight tension compensation through a simple control method, while more advanced schemes are left for future work. For simplicity, index $i$ is omitted hereafter; the outer loop runs at $100$~Hz as the pose estimate on PX4.
\vspace{-0.2cm}

\subsection{Position Control}\label{position_ctrl_cascade}
The outermost block regulates position and provides velocity setpoints through a Proportional controller:
\begin{equation}
    \dot{\bm{\xi}}_d = K_P(\bm{\xi}_d - \bm{\xi}) \quad ,
\end{equation}
where $K_P$ = diag($k_{P,x}$, $k_{P,y}$, $k_{P,z}$) is the diagonal matrix composed of the proportional gains, and $\bm{\xi}_d$ is generated from a nominal target $\bm{\bar{\xi}}_d$, and adjusted using tension feedback:
\begin{equation}
    \bm{\xi}_d = \bm{\bar{\xi}}_d + K_{T_P}\bm{e}_T + K_{T_I} \int_{0}^{t}\bm{e}_T\,dt - K_{T_D}\frac{d}{dt} \bm{T}_{{f}} \quad , \label{tension_control}
\end{equation}
with $K_{T_P}$, $K_{T_I}$, and $K_{T_D}$ diagonal gain matrices in $\mathbb{R}^{3 \times 3}$. The term $\bm{e}_T = \bm{T}_d - \bm{T}_{f}$ is the tension error with respect to the nominal static tension $\bm{T}_d$, expected when the UAV rests at $\bm{\bar{\xi}}_d$, while $\bm{T}_{f}$ is a lightly low-pass-filtered version of $\bm{T}$.

\subsection{Velocity Control}\label{velocity_ctrl_cascade}
The velocity block receives the commanded velocity vector from the previous stage and computes desired accelerations through a Proportional-Integral-Derivative controller:
\begin{equation}
   \ddot{\bm{\xi}}_d = K_{v_P} (\dot{\bm{\xi}}_d - \dot{\bm{\xi}}) + K_{v_I} \int_{0}^{t}(\dot{\bm{\xi}}_d - \dot{\bm{\xi}})\,dt - K_{v_D}\frac{d}{dt} \dot{\bm{\xi}}_{{f}} \quad ,\label{pid_xi}
\end{equation}
where $\dot{\bm{\xi}}_{{f}}$ is the low-pass filtered velocity vector state.
Again, $K_{v_P}, K_{v_I}, K_{v_D}$ are three diagonal gain matrices $\in \mathbb{R}^{3 \times 3}$.

\subsection{Thrust and Attitude Mapping}\label{intermediate_block_cascade}
The resulting acceleration commands, along with the desired yaw $\psi_d$, are mapped into thrust and attitude references for the low-level loops. From the desired force vector:
\begin{equation}
    \bm{F}_d = [F_{d,x} \; F_{d,y} \; F_{d,z}]^{\top} = m \ddot{\bm{\xi}}_d - \bm{P} \quad ,
\end{equation}
the thrust input command is obtained from the $z$-component of the force, corrected by the vehicle tilt:
\begin{equation}
u = F_{d,z}/(c_\phi c_\theta) \footnotemark \quad .
\end{equation}
\footnotetext{$c_{\alpha} = \cos(\alpha)$, $s_{\alpha} = \sin(\alpha)$ for any angle $\alpha$.}
To generate the desired attitude quaternion $\bm{q}_d$, desired axes are defined in $\mathcal{F}_I$: the $z$-axis aligns opposite to the normalized desired force $\tilde{\bm{z}}_d = -\bm{F}_d/||\bm{F}_d||$. By defining a heading vector encoding the yaw reference $\bm{\chi}_{\psi} = [c_{\psi_d} \; s_{\psi_d} \; 0]^\top$, the $y$ and $x$ axes follow from orthogonalization:  
\begin{equation}
    \tilde{\bm{y}}_d = (\tilde{\bm{z}}_d \times \bm{\chi}_{\psi})/(||\tilde{\bm{z}}_d \times \bm{\chi}_{\psi}||)\;, \quad
    \tilde{\bm{x}}_d = \tilde{\bm{y}}_d \times \tilde{\bm{z}}_d \quad .
\end{equation}
Thus, the desired rotation matrix is $R_d = [\tilde{\bm{x}}_d ; \tilde{\bm{y}}_d ; \tilde{\bm{z}}_d]$, from which the quaternion $\bm{q}_d \in \mathbb{R}^4$ is extracted.

\section{Experimental Setup}\label{experimental_setup}
\subsection{Software Architecture} \label{sofar}
The framework used for both \textit{Software-In-The-Loop} simulations and indoor experiments relies on a ROS2 stack, Humble on Ubuntu 22, running high-level control and planning nodes. PX4 v1.16.1, an open-source flight-control software for multirotors and other unmanned vehicles, handles low-level control loops and actuator commands. Gazebo, a physics engine providing realistic dynamics and sensor data, is used only in simulation; it also provides the 3D tension at the UAV-cable joint. QGroundControl (QGC) is used as ground-station software for telemetry, mission planning, and pre-flight parameter tuning, interfacing with PX4 through MAVLink. Communication between ROS2 and PX4 relies on the lightweight uXRCE-DDS protocol, preserving the same software structure in simulation and experiments.

The system behavior is handled by a decentralized state machine running on both UAVs. State transitions are synchronized through ROS2 topics over the same Wi-Fi network, where the UAVs publish ready signals.

\begin{figure*}[t]
\vspace{0.2cm}
\centerline{\includegraphics[scale=0.145]{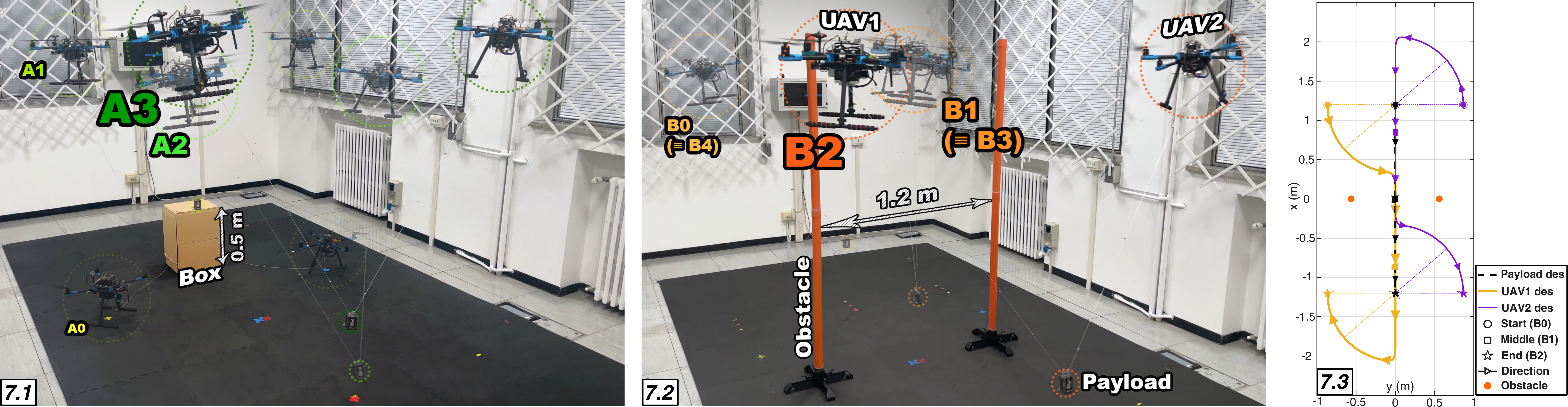}}
\caption{Motion composites of Test A payload-drop (7.1) and Test B constrained-space maneuvers (7.2), with A0-A3 and B0-B4 marking successive instants. Test B top-view desired trajectories are shown in 7.3; return paths are mirrored.}
\vspace{-0.5cm}
\label{system_force_sensor_tests}
\end{figure*}

\vspace{-0.1cm}
\subsection{Hardware Architecture} \label{harar}
The experimental platform consists of two modified \textit{Holybro x500 v2} quadrotors ($\sim2$~kg each, battery included), each equipped with a \textit{Pixhawk 6C} running PX4 and integrating an IMU, barometer, and magnetometer. A 3D-printed PLA battery compartment moves the battery closer to the UAV COM and leaves space for the force sensor. A \textit{Raspberry Pi 4} (8 GB RAM, quad-core Cortex-A72 at 1.8 GHz) runs Ubuntu with ROS2, executes high-level nodes, and communicates with the Pixhawk through a serial uXRCE-DDS link.

External pose feedback is provided by an \textit{OptiTrack} system with eight Flex 13 cameras (Fig.~\ref{system_force_sensor}) covering about $3 \times 4 \times 2.5$~m${}^3$. Measurements are streamed over Wi-Fi to the UAVs and fused on the Pixhawk for odometry. QGC connects to the Pixhawk through MAVLink radio for supervision.

\vspace{-0.1cm}
\subsection{System Evaluation}
The two UAVs are connected to a $1$~kg payload (two $0.5$~kg lead ballasts equipped with MoCap markers for tracking) using $\sim1.5$~m nylon cables and placed at predefined locations and yaw angles ($60$~deg. phase difference between the UAVs) in the MoCap arena.

An experimental campaign is organized into payload-drop tests and narrow-passage maneuvers, assessing sudden load variations, payload oscillations, and dynamic constrained motion. The distributed geometric controller in \cite{SunDario} is used as a simulation reference, while indoor experiments compare the proposed force sensor with tensions derived from the distributed external-force estimator in \cite{SunDario}, using PX4 accelerometer data and motor angular velocities, available on average at $900$~Hz and $60$~Hz, respectively. Consistent with the proposed architecture, all methods rely on limited inter-UAV communication, no payload-state knowledge, and cable tensions either measured onboard or estimated locally.

Robustness to uncertain position feedback is evaluated by artificially perturbing the UAV position with white Gaussian noise $\bm{w}_i(t)\sim\mathcal{N}(\bm{0},\Sigma_i)$ and a slowly varying bias $\bm{b}_i(t)$ modeled as a first-order Gauss-Markov process $\dot{\bm{b}}_i(t) = -\bm{b}_i(t)/{\tau_b} + \bm{\eta}_i(t)$, with $\bm{\eta}_i(t)\sim\mathcal{N}(\bm{0},Q_i)$ and correlation time $\tau_b>0$. Different random seeds are used for the two UAVs to ensure independent noise sequences.

The parameters of our controller are defined as: $K_P$ = diag(0.8, 0.8, 0.7), $K_{v_P}$ = diag(2.7, 2.7, 6.0), $K_{v_I}$ = diag(1.9, 1.9, 3.0), $K_{v_D}$ = diag(0.3, 0.3, 0.5)), $K_{T_P}$ = diag(0.22, 0.22, 0.06), $K_{T_I}$ = diag(0.08, 0.08, 0.18), $K_{T_D}$ = diag(0.02, 0.02, 0.0). Disturbance parameters are: $\tau_b = 60.0$~s, $\Sigma_i$ = diag(0.03, 0.03, 0.06),  $Q_i$ = diag(0.08, 0.08, 0.15).

\section{Results} \label{results}

Several trials were conducted; representative cases are shown next.
Flying phases are \textit{Takeoff} and \textit{Move}, with transitions triggered when both UAVs stay within $0.3$~m of their position setpoints for $1$~s. The graphs report UAV and payload states as detected by the MoCap system; the desired setpoints are shown as the nominal values $\bm{\bar{\xi}}_d$.

\vspace{-0.1cm}
\subsection*{Test A - Payload Drop}
In these tests, the payload is initially placed on a box at $\bm{\xi}_p(0) = [1.0, 0.0, -0.5]^\top$ (Fig.~\ref{system_force_sensor_tests}.1, frame A0). The two UAVs take off to the desired position with slack cables and, once ready (A1), translate by $2$~m toward a new setpoint, causing the payload to drop from the box (A2). The resulting sudden cable forces and payload oscillations are then stabilized in the final phase (A3).

\begin{figure}[!b]
\vspace{-0.25cm}
\centerline{\includegraphics[scale=0.41]{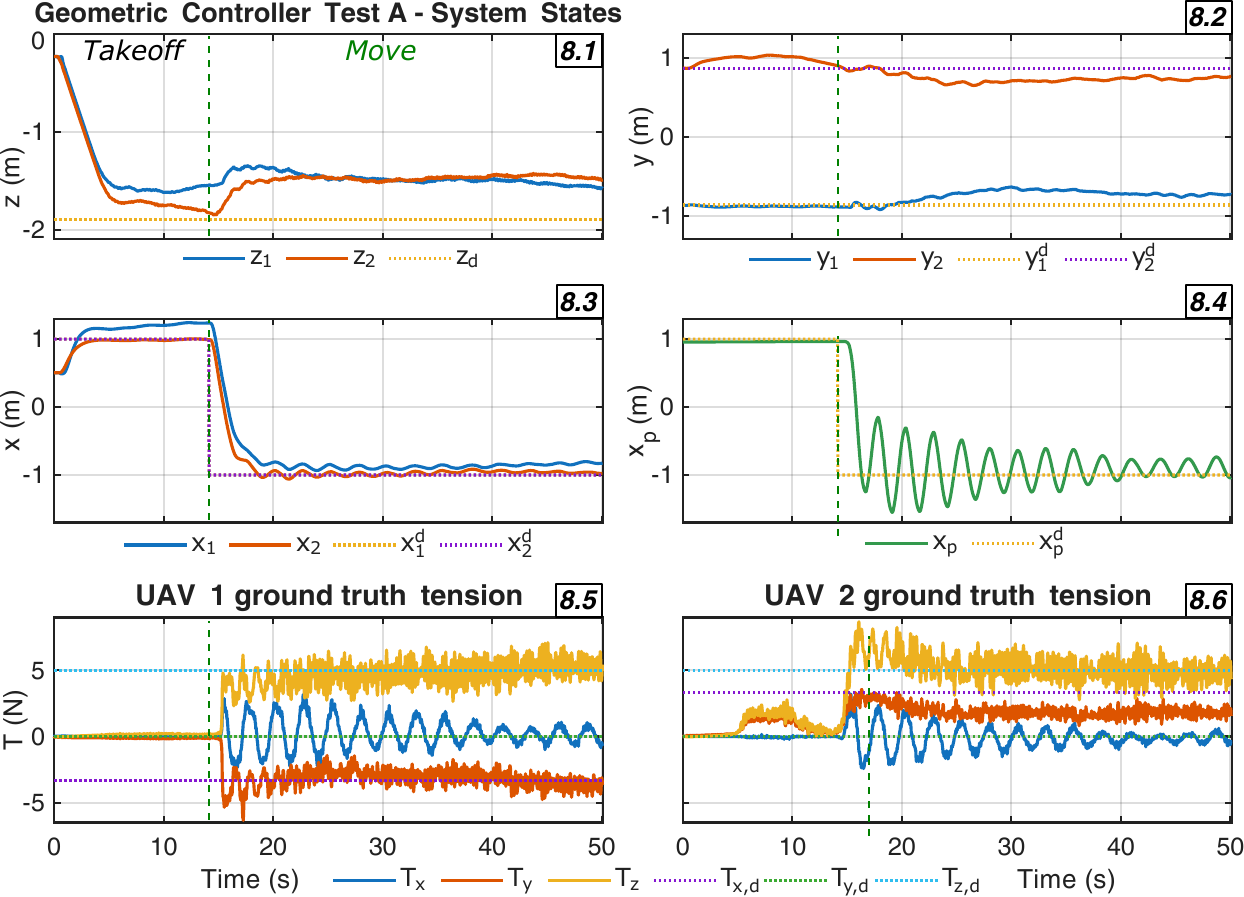}}
\caption{Simulation results of system states in test A with geometric controller. Legend is shared for measured tensions.
}

\vspace{-0.3cm}
\label{all_in_one_sim}
\end{figure}

\subsubsection{Simulation}

The distributed geometric controller in \cite{SunDario} is first tested in simulation using exact cable tensions from a Gazebo plugin, to isolate the control strategy. As shown in Fig.~\ref{all_in_one_sim}, it performs well in \textit{Takeoff}, with both UAVs reaching the state-change condition, despite biased positions and minor tension oscillations on UAV2 as its cable approaches the taut condition. After the payload drop in \textit{Move} phase, it initially reduces the $y$ deviation through direct tension compensation (Fig.~\ref{all_in_one_sim}.2).

However, during stabilization, the $z$ setpoint is not tracked (Fig.~\ref{all_in_one_sim}.1), and oscillations appear, likely due to direct compensation under position uncertainty. This is visible in the $x$ displacement, tensions, and especially payload motion, where oscillations persist at steady state (Fig.~\ref{all_in_one_sim}.3-6). To avoid this oscillatory response, experiments use the proposed controller and compare tensions derived from the external-force estimator in \cite{SunDario} with force-sensor measurements.

\subsubsection{Experiment}
\begin{figure}[t]
\vspace{0.2cm}
\centerline{\includegraphics[scale=0.41]{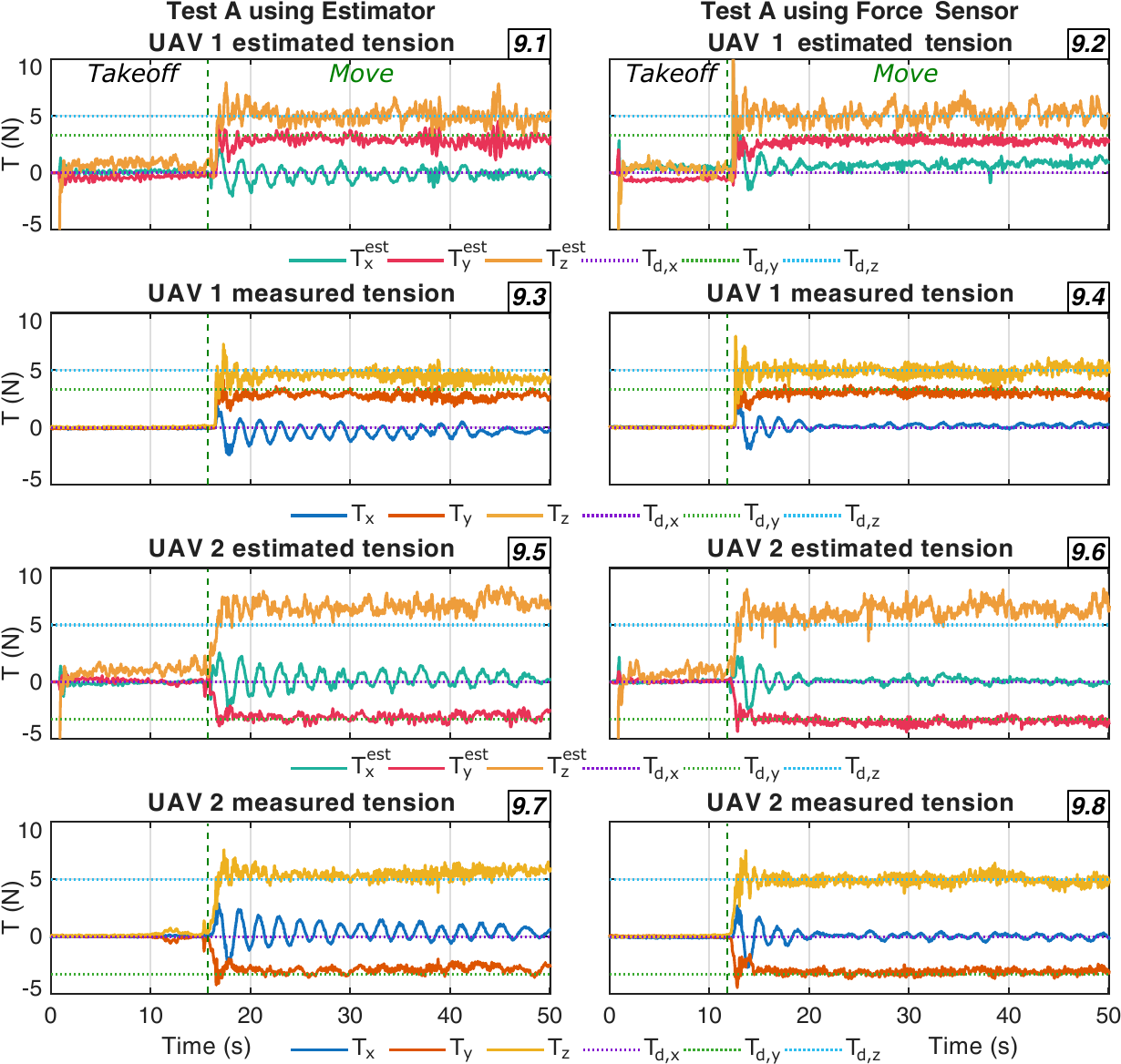}}
\caption{Experimental results of estimated and measured tensions in test A using Estimator (left column) and Force Sensor (right column).\protect\commonfignote}
\vspace{-0.4cm}
\label{tensions_all_exp}
\end{figure}

The indoor experiments are first analyzed by comparing estimated and measured tensions in Fig.~\ref{tensions_all_exp}. Before \textit{Takeoff}, the estimator shows a spike since gravity is not yet balanced by thrust. During \textit{Takeoff}, it deviates from the force sensor, mainly showing an almost constant bias in the $z$ component for both UAVs and in the $y$ component for UAV1, together with chattering likely due to thrust estimation and amplified by position uncertainty.

After the payload drop, the estimator follows the force sensor more closely: the $z$ component agrees well during the first $\sim4$~s of \textit{Move}, while the $x$-$y$ components remain consistent throughout the phase. However, the biases reappear, becoming more evident for UAV2 despite using the same motor constant for both vehicles. Other discrepancies include isolated outliers, such as the $z$ estimate reaching $\sim27$~N at the beginning of \textit{Move} in Fig.~\ref{tensions_all_exp}.2 (outside the plot range), downward spikes in Fig.~\ref{tensions_all_exp}.6, and noise intervals in Fig.~\ref{tensions_all_exp}.1 from $\sim37$~s onward, bounded or absent in the corresponding measured tension (Fig.~\ref{tensions_all_exp}.3).

\begin{figure}[t]
\vspace{0.2cm}
\centerline{\includegraphics[scale=0.41]{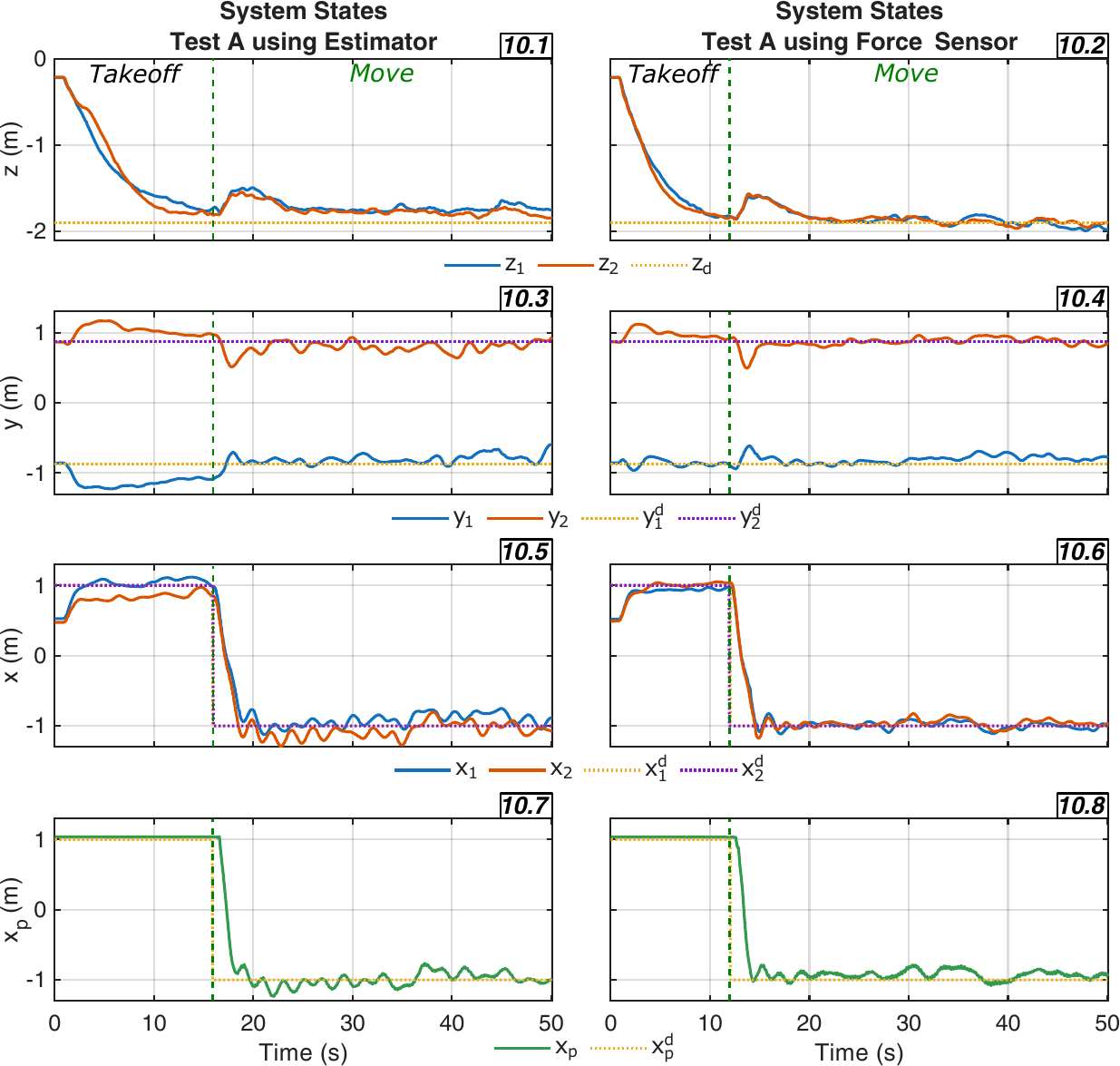}}
\caption{Experimental results of system states in test A using Estimator (left column) and Force Sensor (right column).\protect\commonfignote}
\vspace{-0.3cm}
\label{all_in_one_exp}
\end{figure}

Overall, the force sensor provides cleaner and more reliable tension data while capturing small force variations. For instance, around $12$~s in Fig.~\ref{tensions_all_exp}.7, just before the \textit{Move} phase, the UAV2 cable slightly tightens and a small tension is measured, unlike with the estimator in Fig.~\ref{tensions_all_exp}.5. Although the estimator in \cite{SunDario} remains lightweight when approximate tension information is sufficient, e.g., to keep tensions within planner-defined bounds, measured tensions provide cleaner feedback and improve UAV-payload stability, especially in the $x$-motion after the drop (Fig.~\ref{all_in_one_exp}.6, \ref{all_in_one_exp}.8).

\subsection*{Test B - Narrow Passage Maneuvers}
The final experiment evaluates a dynamic narrow-passage maneuver using only the proposed methods, since payload-drop tests showed cleaner feedback and better stabilization with measured tensions. Starting from hover with taut cables (Fig.~\ref{system_force_sensor_tests}.2, frame B0, the UAVs change formation to cross two obstacles in the MoCap room (B1), recover it (B2), wait $1$~s, and perform the reverse maneuver (B3) to return to the start (B4). Formation changes use circular trajectories, while obstacle crossing uses translational ones (check desired top-view trajectories in Fig.~\ref{system_force_sensor_tests}.3); setpoints define an $11$~s round trip, with two $5$~s passages separated by the $1$~s wait.

\begin{figure}[t]
\vspace{0.2cm}
\centerline{\includegraphics[scale=0.41]{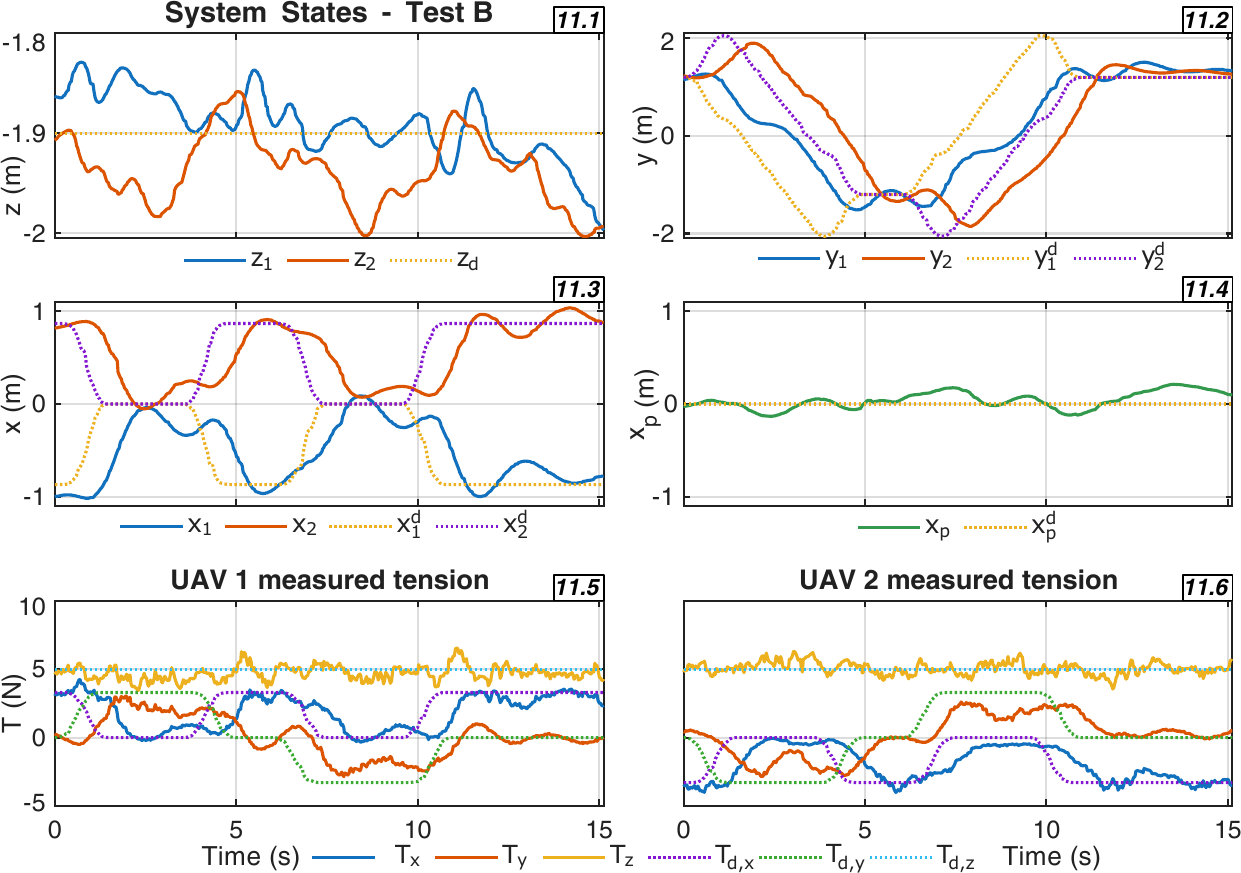}}
\caption{Experimental results of system states in test B. Legend is shared for measured tensions.}
\vspace{-0.5cm}
\label{narrow_passage_test}
\end{figure}

Performances during \textit{Move} are shown in Fig.~\ref{narrow_passage_test}. Despite the limited maneuver time, the overall system response remains smooth. Along $z$, the UAVs regulate their altitude to keep the measured tensions close to the desired values (Fig.~\ref{narrow_passage_test}.1,\ref{narrow_passage_test}.5-6). Along $x$-$y$, tracking degrades, likely due to inertia and conservative control gains kept unchanged across tests. This is most evident for UAV1 along $y$ during obstacle crossing (Fig.~\ref{narrow_passage_test}.2, \ref{narrow_passage_test}.5-6, intervals $1.3$-$3.7$~s and $7.3$-$9.7$~s), also causing larger deviations in the tension $y$ component for both UAVs. In the remaining intervals, however, the system robustly tracks the desired tensions.

A supplementary video is available at: \url{https://youtu.be/rIw9-fvV8Qw} to offer clearer visual feedback on the experimental results.

\begingroup
\renewcommand{\thefootnote}{$\S$}
\footnotetext{\hypertarget{fn:commonfig}{}Rows compare the same quantities across tests; legends are shared.}
\endgroup

\section{Conclusions} \label{Conclusions}
This work proposed a dual-UAV payload transportation framework using a custom force sensor for direct cable-tension measurement. The sensor was designed, modeled, and characterized. A decentralized cascade-control architecture exploiting force feedback has been presented. Simulations and indoor experiments under position uncertainty validated the approach through payload-drop and narrow-passage tests against methods from the literature. Results showed improved stability, coordination, and disturbance rejection with the proposed sensor and controller.

Future work will address both hardware improvements and more advanced decentralized control strategies for cooperative transport. The force sensor can be further optimized by reducing the aluminium support weight, compensating temperature drift, increasing data rate, defining operating limits, and improving resistance to dust and splashes.


\vspace{-0.05cm}

\begin{thebibliography}{00}

\bibitem{CaccavaleRobArms}
F. Caccavale, G. Giglio, G. Muscio, and F. Pierri, ``Cooperative impedance control for multiple UAVs with a robotic arm,'' in \emph{Proc. IEEE/RSJ Int. Conf. Intell. Robots Syst. (IROS)}, Hamburg, Germany, 2015, pp. 2366-2371.

\bibitem{Tagliabue}
A. Tagliabue, M. Kamel, R. Siegwart, and J. Nieto, ``Robust collaborative object transportation using multiple MAVs,'' \emph{Int. J. Robot. Res.}, vol. 38, no. 9, pp. 1020-1044, 2019.

\bibitem{Loianno}
G. Loianno and V. Kumar, ``Cooperative transportation using small quadrotors using monocular vision and inertial sensing,'' \emph{IEEE Robot. Autom. Lett.}, vol. 3, no. 2, pp. 680-687, 2018.

\bibitem{Bosio}
C. Bosio and M. W. Mueller, ``Automated layout and control co-design of robust multi-UAV transportation systems,'' \emph{IEEE Robot. Autom. Lett.}, vol. 10, no. 4, pp. 3956-3963, Apr. 2025.

\bibitem{DarioFirstPP}
D. Sanalitro, H. J. Savino, M. Tognon, J. Cortés, and A. Franchi, ``Full-pose manipulation control of a cable-suspended load with multiple UAVs under uncertainties,'' \emph{IEEE Robot. Autom. Lett.}, vol. 5, no. 2, pp. 2185-2191, Apr. 2020.

\bibitem{PickPlaceDario}
A. E. Jiménez-Cano, D. Sanalitro, M. Tognon, A. Franchi, and J. Cortés, ``Precise cable-suspended pick-and-place with an aerial multi-robot system,'' \emph{J. Intell. Robot. Syst.}, vol. 105, no. 3, Jul. 2022.

\bibitem{Jackson}
B. E. Jackson, T. A. Howell, K. Shah, M. Schwager, and Z. Manchester, ``Scalable cooperative transport of cable-suspended loads with UAVs using distributed trajectory optimization,'' \emph{IEEE Robot. Autom. Lett.}, vol. 5, no. 2, pp. 3368-3374, Apr. 2020.

\bibitem{MPCSundin}
R. C. Sundin, P. Roque, and D. V. Dimarogonas, ``Decentralized model predictive control for equilibrium-based collaborative UAV bar transportation,'' in \emph{Proc. IEEE Int. Conf. Robot. Autom. (ICRA)}, Philadelphia, PA, USA, 2022, pp. 4915-4921.

\bibitem{Buzzurro}
N. De Carli, R. Belletti, E. Buzzurro, A. Testa, G. Notarstefano, and M. Tognon, ``Distributed NMPC for cooperative aerial manipulation of cable-suspended loads,'' \emph{IEEE Robot. Autom. Lett.}, vol. 10, no. 10, pp. 10546-10553, Oct. 2025.

\bibitem{Scaramuzza}
M. Gassner, T. Cieslewski, and D. Scaramuzza, ``Dynamic collaboration without communication: Vision-based cable-suspended load transport with two quadrotors,'' in \emph{Proc. IEEE Int. Conf. Robot. Autom. (ICRA)}, 2017, pp. 5196-5202.

\bibitem{XieLasVegas}
H. Xie, X. Cai, and P. Chirarattananon, ``Towards cooperative transport of a suspended payload via two aerial robots with inertial sensing,'' in \emph{Proc. IEEE/RSJ Int. Conf. Intell. Robots Syst. (IROS)}, Las Vegas, NV, USA, 2020, pp. 1215-1221.

\bibitem{GabellieriForceCtrl}
C. Gabellieri, M. Tognon, D. Sanalitro, and A. Franchi, ``Force-based pose regulation of a cable-suspended load using UAVs with force bias,'' in \emph{Proc. IEEE/RSJ Int. Conf. Intell. Robots Syst. (IROS)}, Detroit, MI, USA, 2023, pp. 6920-6926.

\bibitem{TognonIF}
M. Tognon, C. Gabellieri, L. Pallottino, and A. Franchi, ``Aerial co-manipulation with cables: The role of internal force for equilibria, stability, and passivity,'' \emph{IEEE Robot. Autom. Lett.}, vol. 3, no. 3, pp. 2577-2583, Jul. 2018.

\bibitem{TagliabueOptSens}
A. Tagliabue, M. Kamel, S. Verling, R. Siegwart, and J. Nieto, ``Collaborative transportation using MAVs via passive force control,'' in \emph{Proc. IEEE Int. Conf. Robot. Autom. (ICRA)}, Singapore, 2017, pp. 5766-5773.

\bibitem{ZhangRAL}
X. Zhang, F. Zhang, P. Huang, J. Gao, H. Yu, and C. Pei, ``Self-triggered based coordinate control with low communication for tethered multi-UAV collaborative transportation,'' \emph{IEEE Robot. Autom. Lett.}, vol. 6, no. 2, pp. 1559-1566, Apr. 2021.

\bibitem{PereiraTensionState}
P. O. Pereira, P. Roque, and D. V. Dimarogonas, ``Asymmetric collaborative bar stabilization tethered to two heterogeneous aerial vehicles,'' in \emph{Proc. IEEE Int. Conf. Robot. Autom. (ICRA)}, Brisbane, QLD, Australia, 2018, pp. 5247-5253.

\bibitem{HRCollabTransp}
G. Li, X. Liu, and G. Loianno, ``Human-aware physical human-robot collaborative transportation and manipulation with multiple aerial robots,'' \emph{IEEE Trans. Robot.}, vol. 41, pp. 762-781, 2025.

\bibitem{DelbeneTension}
A. Delbene and M. Baglietto, ``Cables tension modeling for multi-UAV payload transportation,'' in \emph{Proc. IEEE Int. Conf. Autom. Sci. Eng. (CASE)}, Bari, Italy, 2024, pp. 212-218.

\bibitem{SunDario}
S. Sun, X. Wang, D. Sanalitro, A. Franchi, M. Tognon, and J. Alonso-Mora, ``Agile and cooperative aerial manipulation of a cable-suspended load,'' \emph{Science Robot.}, vol. 10, no. 107, Oct. 2025.

\bibitem{GabellieriCyclicNullspace}
C. Gabellieri, Y. Shen, M. Paolucci, and A. Franchi, ``Cyclic nullspace coordination: Perpetual flight of aerial carriers for static suspension,'' \emph{IEEE Trans. Control Syst. Technol.}

\bibitem{GimenezUltra}
L. R. Salinas, J. Gimenez, D. C. Gandolfo, C. D. Rosales, and R. Carelli, ``Unified motion control for multilift unmanned rotorcraft systems in forward flight,'' \emph{IEEE Trans. Control Syst. Technol.}, vol. 31, no. 4, pp. 1607-1621, 2023.

\bibitem{WahbaIROS24}
K. Wahba, J. Ortiz-Haro, M. Toussaint, and W. Hönig, ``Kinodynamic motion planning for a team of multirotors transporting a cable-suspended payload in cluttered environments,'' in \emph{Proc. IEEE/RSJ Int. Conf. Intell. Robots Syst. (IROS)}, Abu Dhabi, United Arab Emirates, 2024, pp. 12750-12757.

\bibitem{BendingPayload}
J. Xu, L. Gao, R. Fierro, and D. Saldaña, ``AirBender: Adaptive transportation of bendable objects using dual UAVs,'' \emph{IEEE Robot. Autom. Lett.}, vol. 10, no. 3, pp. 2790-2797, Mar. 2025.

\bibitem{FlyCrane}
D. Sanalitro, M. Tognon, A. E. Jiménez-Cano, J. Cortés, and A. Franchi, ``Indirect force control of a cable-suspended aerial multi-robot manipulator,'' \emph{IEEE Robot. Autom. Lett.}, vol. 7, no. 3, pp. 6726-6733, Jul. 2022.

\bibitem{EnergyDistr}
A. Mohiuddin, Y. Zweiri, R. Almadhoun, T. Taha, and D. Gan, ``Energy distribution in dual-UAV collaborative transportation through load sharing,'' \emph{J. Mechanisms Robot.}, pp. 1-14, Apr. 2020.

\bibitem{BernardForceSens1}
M. Bernard, K. Kondak, I. Maza, and A. Ollero, ``Autonomous transportation and deployment with aerial robots for search and rescue missions,'' \emph{J. Field Robot.}, vol. 28, pp. 914-931, 2011.

\bibitem{BernardForceSens2}
M. Bernard and K. Kondak, ``Generic slung load transportation system using small size helicopters,'' in \emph{Proc. IEEE Int. Conf. Robot. Autom. (ICRA)}, Kobe, Japan, 2009, pp. 3258-3264.

\bibitem{SRIM37}
H. Li, H. Zhong, J. Gao, Y. Lv, J. Sha, J. Liang, and H. Zhang, ``A nonlinear trajectory tracking control strategy for quadrotor with suspended payload based on force sensor,'' \emph{IEEE Trans. Intell. Veh.}, vol. 9, no. 1, pp. 704-714, 2024.

\bibitem{DelbeneRecovery}
A. Delbene and M. Baglietto, ``Recovery techniques for a multi-UAV system transporting a payload,'' in \emph{Proc. IEEE Int. Conf. Autom. Sci. Eng. (CASE)}, Bari, Italy, 2024, pp. 559-566.

\bibitem{PX4_ICRA}
L. Meier, D. Honegger, and M. Pollefeys, ``PX4: A node-based multithreaded open source robotics framework for deeply embedded platforms,'' in \emph{Proc. IEEE Int. Conf. Robot. Autom. (ICRA)}, Seattle, WA, USA, 2015, pp. 6235-6240.

\end{thebibliography}
\end{document}